\documentclass[10pt,twocolumn,letterpaper]{article}
\PassOptionsToPackage{table}{xcolor}

\usepackage{cvpr}              
\definecolor{cvprblue}{rgb}{0.21,0.49,0.74}
\usepackage[pagebackref,breaklinks,colorlinks,allcolors=cvprblue]{hyperref}

\def\confName{CVPR}
\def\confYear{2027}

\usepackage{hyperref}       
\usepackage{url}            
\usepackage{booktabs}  
\usepackage{xcolor}  
\usepackage{cancel}
\usepackage{amsmath}
\usepackage{amssymb}
\usepackage{bbm}
\usepackage{dsfont}
\usepackage{multirow}
\usepackage{enumitem}
\usepackage{colortbl}
\usepackage{wrapfig}
\usepackage[most]{tcolorbox}
\newtcolorbox{takeaway}{
  colback=blue!3,     
  colframe=blue!70!black, 
  arc=2mm,            
  boxrule=0.8pt       
}
\usepackage{arydshln}
\usepackage{algorithm}
\usepackage{algpseudocode}
\usepackage{listings}
\usepackage{ulem}
\definecolor{beige}{RGB}{250,240,210}
\definecolor{peach}{RGB}{255,220,180}
\definecolor{deeppeach}{RGB}{200,115,70}
\definecolor{lightpink}{RGB}{255, 235, 240}
\definecolor{mint}{RGB}{225,245,237}
\definecolor{lightpurple}{RGB}{235,210,255} 
\definecolor{purple}{RGB}{145,115,180}
\definecolor{softblue}{RGB}{0, 160, 220}
\definecolor{softgreen}{RGB}{90,150,125}
\definecolor{deepgreen}{RGB}{35, 92, 96}
\definecolor{midgray}{gray}{0.65}
\definecolor{lightgray}{gray}{0.9}
\definecolor{deepred}{HTML}{D7191C} 
\definecolor{darkred}{RGB}{192, 0, 0}
\usepackage{wrapfig}
\newcommand{\cc}{\textbf{ProSE}}
\newcommand{\ver}[1]{\textcolor{gray}{$_\mathrm{#1}$}}

\newcommand{\note}[1]{\textcolor[RGB]{196,119,72}{\boldmath{#1}}}

\newcolumntype{Y}{>{\centering\arraybackslash}X}
\usetikzlibrary{arrows.meta}

\newcommand{\uturnarrow}{%
  \tikz[baseline=-0.6ex]\draw[-{Stealth[length=1.6mm]}, line width=1pt, green!70!black]
    (0.35,0.12) -- (0.35,-0.1) -- (0,-0.1) -- (0,0.17);
}
\usepackage[table]{xcolor}

\title{
\cc: Probing Reconstruction Error for Semantic-guided Expert Skipping \\in MoE-MLLMs 
}

\author{
Jiameng Li$^1$~~~~
Minye Wu$^2$~~~~Matthew B. Blaschko$^1$\\
$^1$KU Leuven~~$^2$University of Derby
}

\begin{document}
\maketitle

\begin{abstract}

Mixture-of-Experts Multimodal Large Language Models (MoE-MLLMs) offer scalable inference through sparse expert activation, yet substantial computational redundancy remains among the activated experts.  
Existing expert skipping methods typically calibrate thresholds based on target skipping ratios or final-output divergence, overlooking the accumulation of skipping-induced error across layers and the varying semantic importance of individual tokens. 
To address that, we propose \cc, a training-free, semantic-guided expert skipping framework that probes layer-wise reconstruction error to promote representation consistency compared with default activation. 
Specifically, we construct query-aware semantic masks to identify salient visual tokens and progressively calibrate modality-specific skipping thresholds for each MoE layer. 
By minimizing masked reconstruction error under a given budget, \cc~maintains semantic consistency after skipping and allows adaptive expert allocation under limited activation budgets.
Extensive experiments on 9 benchmarks demonstrate consistent improvements on average over existing expert skipping methods, particularly under aggressive skipping settings. 
\cc~maintains over 96\% performance on KimiVL-A3B-Inst with  83\% expert skipping and accelerates prefilling by $1.7\times$. \textcolor{deeppeach}{Code: }\href{https://github.com/renaissanceee/ProSE}{\textcolor{deeppeach}{https://github.com/renaissanceee/ProSE}}
\end{abstract}    
\section{Introduction}
\label{sec:intro}

Mixture-of-Experts Multimodal Large Language Models (MoE-MLLMs) scale model capacity while controlling computation through sparse expert activation \citep{Qwen3VL,wang2025internvl3_5,li2025minimax,wu2024deepseek}.
For each token, a router selects the Top-K experts and aggregates their outputs according to routing weights.
However, redundancy remains among these activated experts \citep{lu2024not,bai2025diep,huang2026modes}.
Intuitively, when a token's routing weights are concentrated on a few experts, retaining only those experts may suffice, suggesting opportunities to reduce expert computation further \citep{chen2026you}.
Expert skipping exploits this redundancy by selectively bypassing low-contribution experts and preserving model performance. The central challenge is to determine which experts to skip under a given computational budget while minimizing performance degradation.

\begin{figure}[t]
\centering
\begin{subfigure}[h]{0.47\linewidth}
    \centering
    \includegraphics[width=\linewidth]{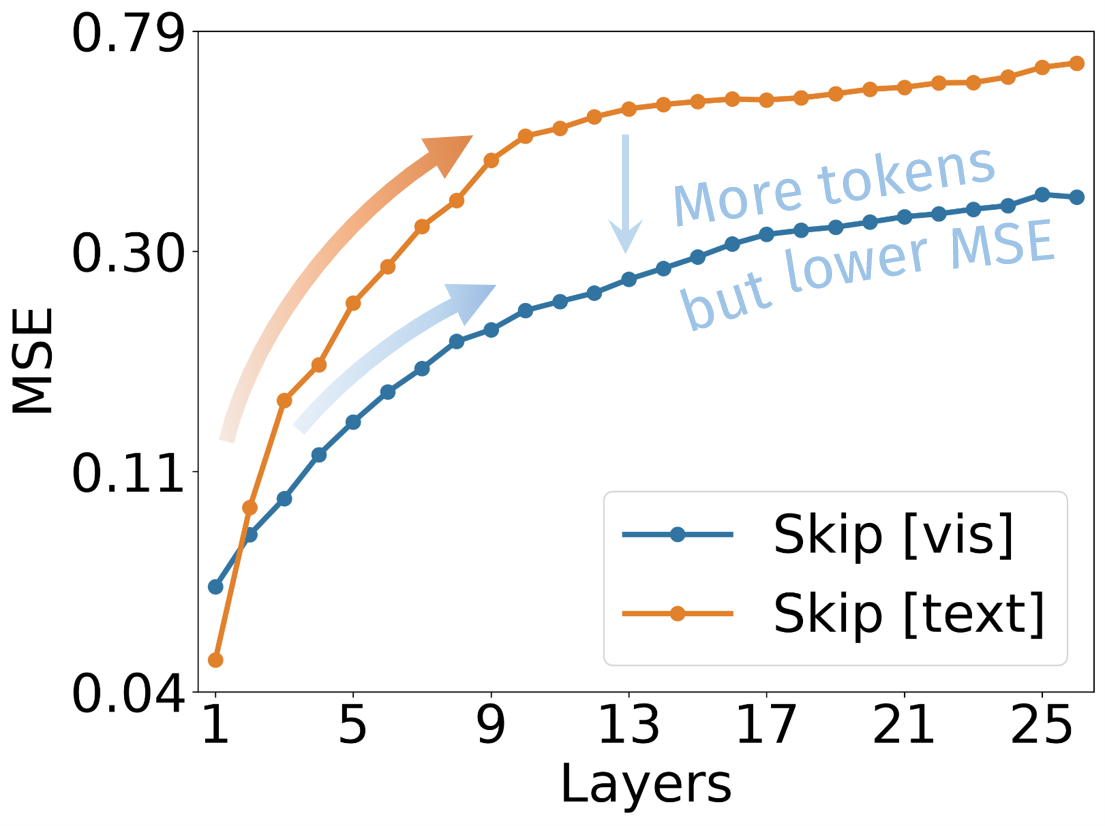}
\end{subfigure}
\begin{subfigure}[h]{0.5\linewidth}
    \centering
    \includegraphics[width=\linewidth]{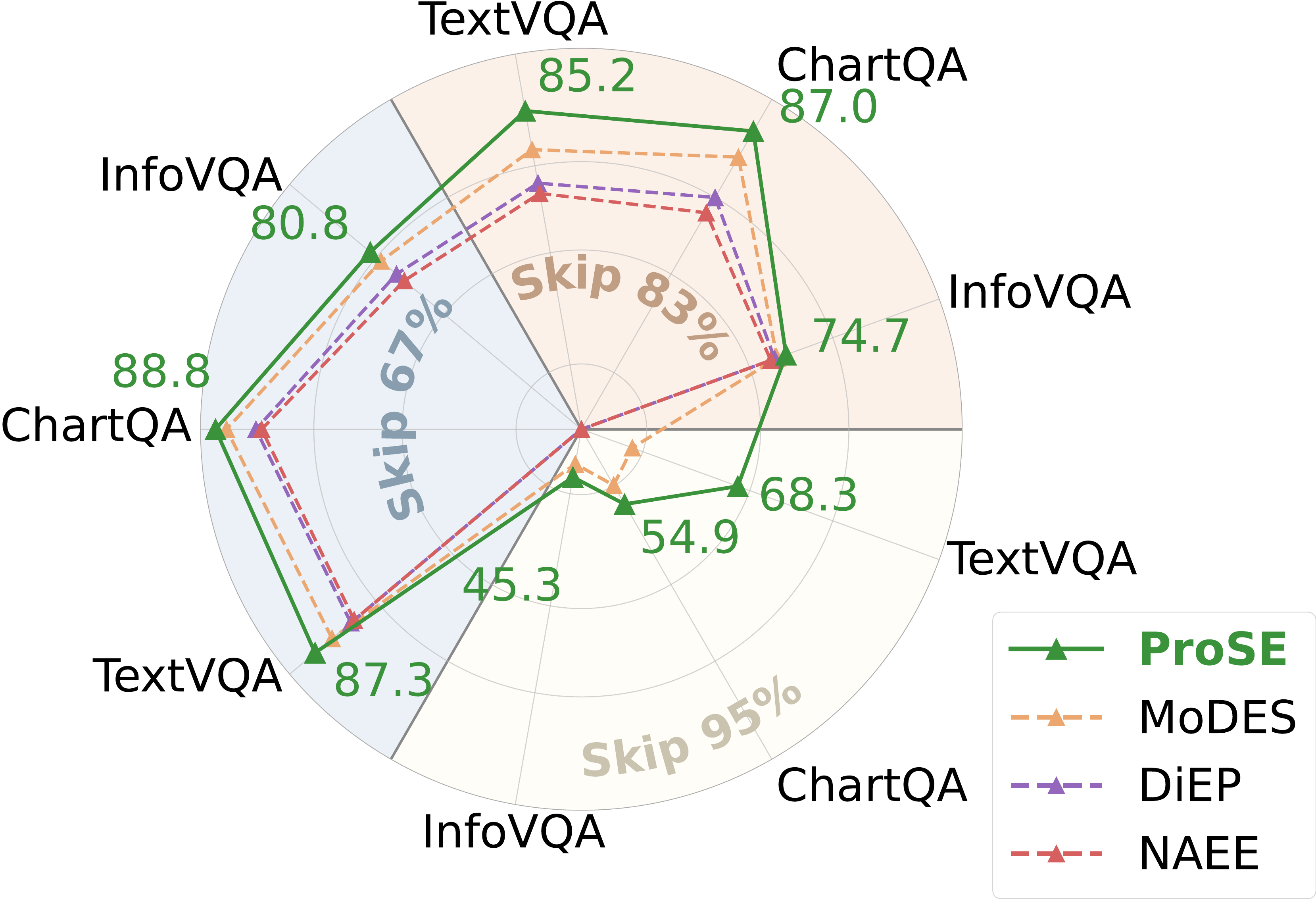}
\end{subfigure}
\caption{
\textbf{Illustration on KimiVL-A3B-Inst \citep{kimiteam2025kimivl}.}
\textbf{(Left)} We skip experts for vision/text tokens separately and extend skipping depths to observe the final-layer token MSE.
Error accumulates on both curves, and skipping for vision tokens yields lower distortion despite their greater number, showing token heterogeneity. 
\textbf{(Right)} ProSE achieves SOTA performance across benchmarks and skipping ratios
(\textcolor[HTML]{6EA6CF}{\textbf{67\%}},
\textcolor[HTML]{D690AF}{\textbf{83\%}},
\textcolor[HTML]{D4BE91}{\textbf{95\%}}).
}
\label{fig:fig1}
\vspace{-0.2cm}
\end{figure}

Existing methods address this challenge using routing information. 
Early approaches, including NAEE \citep{lu2024not} and DiEP \citep{bai2025diep}, primarily investigate unimodal MoE models with Top-2 activation \citep{jiang2024mixtral}.
Modern MoE-MLLMs involve more activated experts and heterogeneous multimodal inputs \citep{Qwen3VL}, requiring more flexible skipping.
MoDES \citep{huang2026modes} adopts a two-stage strategy by estimating layer importance (stage-1) for modulated routing weights and calibrating global visual and textual thresholds (stage-2) using output divergence.
Its design accounts for differences across layers and modalities, enabling dynamic expert allocation in multimodal models.

However, different intermediate skipping behaviors may yield the same output divergence (\textit{ill-posed}), and semantic drift from skipping accumulates across layers (\textit{ill-conditioned}), undermining MoDES's effectiveness.
Moreover, tokens vary in semantic importance \citep{yang2025visionzip,zhang2025cdpruner}, requiring calibration to account for both representation drift and token heterogeneity in the intermediate layers. 
To examine these effects, we retain the Top-1 expert and skip others for visual and textual tokens separately, monitoring the resulting distortion by token-wise MSE at the final layer.
Fig.~\ref{fig:fig1}(left) shows that the representation distortion grows with skipping depth and is lower when skipping experts for visual tokens than for textual tokens.
These observations motivate layer-wise calibration considering token importance.

In this paper, we propose \cc, a training-free expert skipping framework that probes layer-wise reconstruction error to determine the modality-aware thresholds.
Taking the original Top-K model as reference, \cc~calibrates visual and textual thresholds for each MoE layer by minimizing reconstruction error under a target skipping budget.
Considering visual redundancy \citep{yang2025visionzip,zhang2025cdpruner}, we use a query-aware semantic mask to emphasize the representation consistency of informative visual tokens.
The mask only determines which tokens contribute to the calibration objective while not affecting the forward pass.
At inference time, \cc~directly applies the calibrated thresholds for skipping without extra overhead.

Experiments across 9 benchmarks demonstrate improved performance over existing expert skipping methods, with particularly pronounced gains under aggressive settings (\textit{e.g.}, 95\% skipping).
Ablation studies support the effectiveness of layer-wise calibration and show additional benefits from semantic masking.
On KimiVL-A3B-Inst \citep{kimiteam2025kimivl}, \cc~retains 96\% of the original performance at a target skipping ratio of 83\% and achieves $1.7\times$ prefill speedup.
Our contributions are summarized as follows:
\begin{itemize}[nosep,leftmargin=0.4cm]
    \item We propose \cc, a training-free expert skipping approach that determines skipping thresholds by minimizing reconstruction error over important tokens.
    \item Considering token heterogeneity and intermediate semantic drift, we incorporate semantic masks and layer-wise thresholding for fine-grained calibration (Sec.~\ref{sec:method}).
    \item Extensive experiments demonstrate that \cc~improves the trade-off between performance and efficiency over existing skipping methods, with particularly pronounced performance gains under aggressive skipping (Sec.~\ref{sec:exp}).
\end{itemize}

\section{Related Work}
\label{sec:related}

\paragraph{MoE-MLLMs.}
A general MLLM consists of a vision encoder, a tokenizer, a projector for alignment and an LLM decoder.
To achieve efficient inference, many techniques have been proposed, 
including token pruning \citep{yang2025visionzip}, quantization \citep{li2025mbq}, speculative decoding \citep{ji2025specvlm} and distillation \citep{shu2025llava}.
Recently, MoE models \citep{Qwen3VL,kimiteam2025kimivl} have gained popularity for their scaling efficiency, which decouples model capacity from computational cost through sparse expert activation.
Specifically, it splits a dense feed-forward network (FFN) into a router and a few experts, then activates and aggregates the Top-K experts by routing weights \citep{jiang2024mixtral}.
As observed in \cite{zheng2026tackling}, MoE is particularly well-suited to MLLMs, as its sparse activation decomposes representation learning across specialized experts and offers a flexible mechanism for modeling modality imbalance \citep{liu2024deepseek,yang2025qwen3}.
The efficiency challenge for MoE models lies in how to selectively activate experts according to practical budgets.
In particular, \textit{expert skipping} provides a training-free solution to identify the most crucial experts and avoid redundant computation.

\vspace{-0.2cm}
\paragraph{Expert skipping.}

Designed for Top-2 activated MoE-LLMs (\textit{e.g.}, Mixtral-8x7B \citep{jiang2024mixtral}), NAEE \citep{lu2024not} skips the second expert when its routing weight is substantially smaller than the Top-1.
Subsequently,
MC-MoE \citep{huang2025mixture} protects the most salient 2\% tokens and skips experts as NAEE for the remaining tokens.
And DiEP \citep{bai2025diep} modulates the expert importance by Centered Kernel Alignment (CKA).
As a pioneer in expert skipping of MoE-MLLMs, MoDES \citep{huang2026modes} first iteratively masks MoE layers to estimate their importance and then performs a frontier search to determine modality-specific skipping thresholds. However, the calibration relies solely on output divergence in both stages, without explicitly accounting for \textit{semantic drift} and \textit{token heterogeneity} in intermediate layers. In contrast, our method performs layer-wise threshold optimization guided by semantic masks, yielding SOTA performance.
See the appendix for more expert skipping methods.

\begin{figure*}[ht]
\centering
\includegraphics[width=0.95\linewidth]{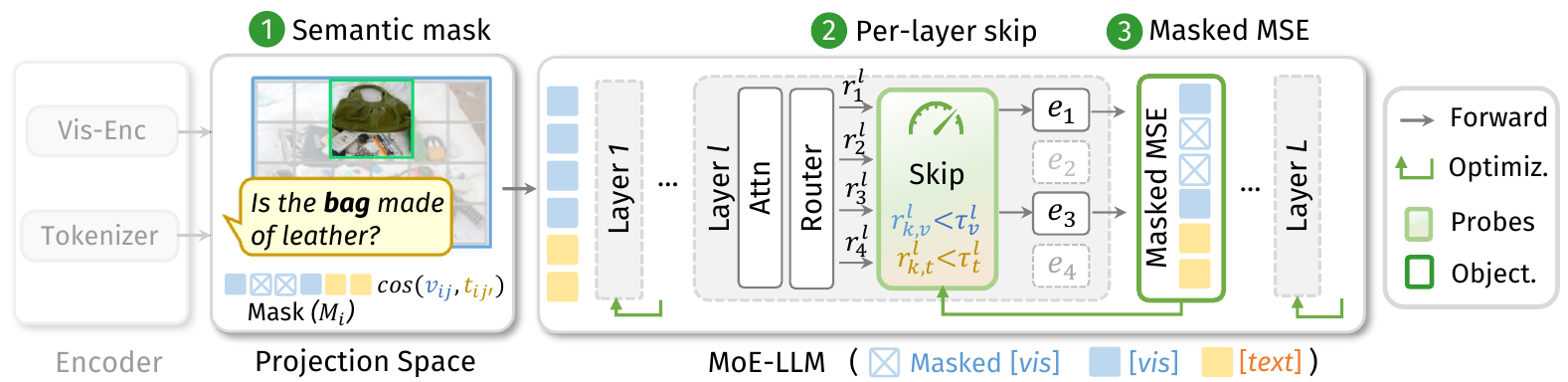}
\caption{
\textbf{Overview of \cc.} During calibration, ProSE constructs a semantic mask ($\mathcal{M}_i$) before LLMs and skips experts in each MoE layer by visual and textual thresholds ($\tau_{v,t}^l$). The thresholds are optimized per layer (\protect\uturnarrow) by minimizing masked MSE. 
}
\label{fig:pipeline}
\end{figure*}

\section{Preliminaries}
\label{subsec:method_pre}

As shown in Fig.~\ref{fig:pipeline}, an MoE layer consists of an attention block, a router and a group of experts (residual connections and normalization omitted). Here, we only visualize $K{=}4$ activated experts for illustration and omit inactivated ones.
A router maps the input token $\mathbf{x}$ to routing probabilities and selects the Top-K experts with corresponding routing weights $\mathbf{r}{=}\{r_1,\ldots,r_K\}$. The output token $\mathbf{y}$ is computed as the weighted sum of the outputs over $K$ activated experts:
\begin{align}
    \mathbf{y}&=\sum_{k=1}^{K} r_k e_k(\mathbf{x}),
    \label{sq:sum}
\end{align}
where $e_k(\cdot)$ denotes the $k$-th activated expert.

To reduce computation, expert skipping methods \citep{huang2026modes,lu2024not,bai2025diep} selectively activate a subset ($\mathcal{S}$) of Top-K experts to calculate the skipped output $\mathbf{\hat{y}}$:
\begin{align}
    \mathbf{\hat{y}} &=\sum_{k \in \mathcal{S} } r_k e_k(\mathbf{x}),~\text{where}~|\mathcal{S}|{<}K.
    \label{sq:sum_skip}
\end{align}
The early work NAEE \citep{lu2024not} identifies $\mathcal{S}$ in MoE-LLMs \citep{jiang2024mixtral} by the Top-2 ratio (${r_2}/{r_1}$), yet struggles to generalize to multimodal settings.
Considering the modality gap in MLLMs, MoDES \citep{huang2026modes} tackles modality-specific subsets $\mathcal{S}{=}\{\mathcal{S}_{v},\mathcal{S}_{t}\}$ for vision and text tokens, which are determined by layer importance ($\alpha^l$) and global vision and text thresholds ($\tau_{v,t}$).
To obtain them, MoDES conducts a layer-wise ablation and a grid search minimizing output perturbation (KL divergence) in succession. 
Despite these designs, MoDES incorporates an inefficient two-stage calibration, and the optimization signal comes far from the end output. 
This coarse supervision also offers limited guidance on which token representations to preserve, given the visual redundancy and token heterogeneity \citep{yang2025visionzip,li2026mi} in MLLMs.  
The oversight of intermediate semantic drift and token importance leads to unsatisfactory performance.

\section{Method}
\label{sec:method}

Our method aims to find effective expert subsets that maintain performance close to default MoE-MLLMs while skipping redundant experts.
To this end, we calibrate layer-wise thresholds separately for visual and textual tokens, considering visual redundancy.
First, we build a semantic token mask based on crossmodal relevance (Sec.~\ref{subsec:method_token}), which ensures the calibration is dominated by salient tokens. 
Expert sets are then selected by applying layer-wise thresholds to the routing weights (Sec.~\ref{subsec:method_threshold}), with modality-specific thresholds for visual and textual tokens.
To determine these thresholds, we conduct layer-wise optimization to minimize the reconstruction error of important tokens within semantic masks (Sec.~\ref{subsec:method_opt}). During inference, \cc~applies thresholds to skip redundant experts without extra overhead (Sec.~\ref{subsec:method_infer}).

\subsection{Semantic Token Mask}
\label{subsec:method_token}

As observed in token pruning methods \citep{yang2025visionzip,zhang2025cdpruner}, visual tokens are inherently redundant with substantial background patches. This redundancy can distract the thresholding, so we introduce a token mask for semantic guidance, allowing the thresholds to be determined by salient tokens. 
As regions of interest are query-dependent \citep{dong2026mmtok}, we define salient tokens as the most query-relevant image patches.
As shown in Fig.~\ref{fig:pipeline}, leveraging the query cues, we measure the visual importance by aggregating maximum crossmodal relevance \citep{li2026mi,dong2026mmtok} in the projection space.
Denote $\mathcal{V}_i{=}\{\mathbf{v}_{ij}\},\mathcal{T}_i{=}\{\mathbf{t}_{ij'}\}$
as visual and textual tokens of sample $i$, where $j,j'$ are the token indices. For each visual token, we compute its pairwise crossmodal relevance and adopt the maximum as its importance 
$s(\mathbf{v}_{ij})$:
\begin{align}
s(\mathbf{v}_{ij})&=\max_{j'} \cos(\mathbf{v}_{ij},\mathbf{t}_{ij'}).
\label{eq:relevance_cross}
\end{align}
Given a visual retention ratio $\beta_v$ (\textit{e.g.} 50\%), we construct a semantic mask $\mathcal{M}_i$ consisting of important vision tokens and all text tokens:
\begin{align}
\mathcal{M}_i &= \operatorname{Top-K}_{\beta_v}(s(\mathcal{V}_i)) \cup \mathbf{1}({\mathcal{T}_i}).
\label{eq:relevance_mask}
\end{align}
$\operatorname{Top-K}_{\beta_v}$ retains top ${\beta_v}$ fraction of visual tokens ranked by crossmodal relevance, and $\mathbf{1}({\mathcal{T}_i})$ selects all text tokens.
Then, the visual and textual tokens are concatenated to form the input to the first LLM layer, denoted as $\mathcal{X}_i^1 {=} \mathcal{V}_i \oplus \mathcal{T}_i$.
Importantly, the mask $\mathcal{M}_i$ selects which tokens contribute to the reconstruction error while not affecting the forward pass, so that the information flow \citep{yang2025vflowopt} is not disrupted.

\begin{figure}[b]
\centering
\begin{subfigure}[h]{\linewidth}
    \centering
    \includegraphics[width=\linewidth]{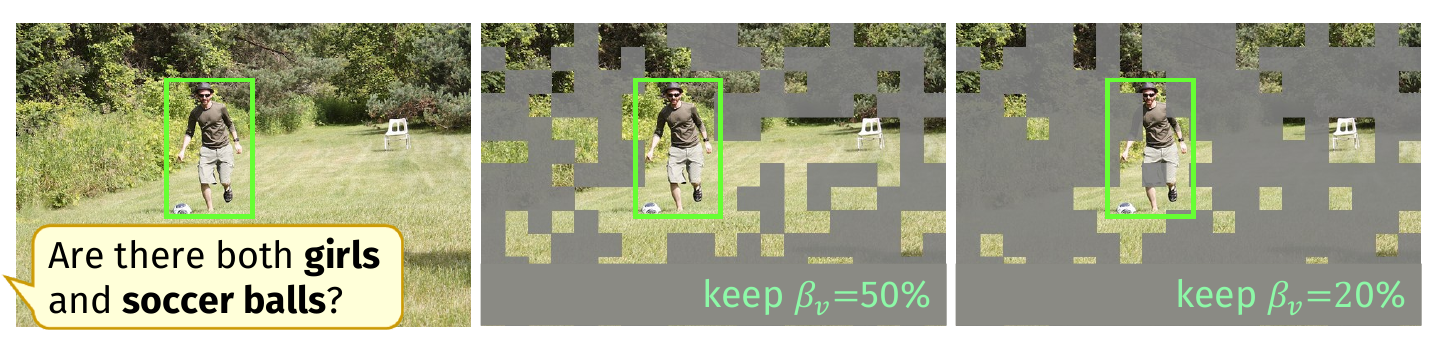}
    \caption{Qwen3VL-MoE-30B-A3B}
    \label{fig:mask_a}
\end{subfigure}
\begin{subfigure}[h]{\linewidth}
    \centering
    \includegraphics[width=\linewidth]{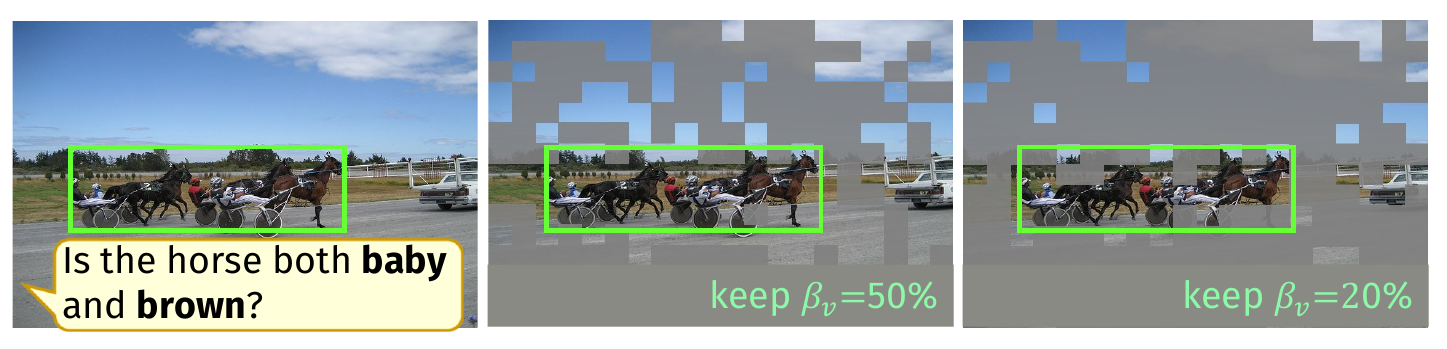}
    \caption{KimiVL-A3B-Inst}
    \label{fig:mask_b}
\end{subfigure}
\vspace{-0.3cm}
\caption{\textbf{Token masks} constructed from GQA \citep{hudson2019gqa}. We highlight key words in \textbf{bold} and query-relevant patches in a \textcolor{softgreen}{\textbf{green}} box. Visual tokens are scored by crossmodal relevance in Eqn.~(\ref{eq:relevance_cross}).
}
\label{fig:mask}
\end{figure}

The token mask plays two roles: (i) reduce the visual proportion in reconstruction error using $\beta_v$, thereby down-weighting visual contributions; (ii) focus the optimization on the semantic consistency of important tokens using $s(\mathcal{V}_i)$. 
As shown in Fig.~\ref{fig:mask}, our query-aware mask effectively captures the most informative patches and excludes most of the background.
We provide results on other importance scores in Sec.~\ref{subsec:exp_ablation}.

\subsection{Layer-wise Thresholding}
\label{subsec:method_threshold}

At layer $l$, we retain experts with routing weights not smaller than $\tau_v^l$ for visual tokens \textcolor{softblue}{[\textit{vis}]} and not smaller than $\tau_t^l$ for textual tokens \textcolor{orange}{[\textit{text}]}: 
\begin{align}
\mathcal{S}=\{k~|~r_{k}^l{\geq}\tau_v^l~\text{for \textcolor{softblue}{[\textit{vis}]},}~\text{or}~r_{k}^l{\geq}\tau_t^l~\text{for \textcolor{orange}{[\textit{text}]}}\}.
\label{eq:rout_ours}
\end{align}
Adopting the selected expert set $\mathcal{S}$, the skipped output 
is calculated as in Eqn.~(\ref{sq:sum_skip}). We visualize our layer-wise thresholds in 
the appendix, where visual ($\tau_{v}^l$) and textual ($\tau_{t}^l$) thresholds generally exhibit substantial gaps, supporting the use of modality-specific thresholds. 
More statistics can be found in our experimental results.
We describe how to obtain the layer-wise dual-thresholds in the following section.

\begin{algorithm}[t]
\caption{\cc}
\label{alg:ours}
\begin{algorithmic}[1]

\Require target skipping ratio $\rho$,
visual ratio $\beta_v$

\Statex \textcolor{midgray}{\textit{// Step 1: Semantic token mask (before LLMs)}}

\State 
$s(\mathbf{v}_{ij}){=}\max_{j'} \cos(\mathbf {v}_{ij},\mathbf{t}_{ij'})$
\Comment{$j,j'$: token index}

\State
$
\mathcal M_{i}{=}\operatorname{Top-K}_{\beta_v}
\bigl(s(\mathcal{V}_i)) \cup \mathbf{1}(\mathcal{T}_i)
$
\Comment{$i$: sample index}

\State
$
\mathcal{X}_i^1{=}\mathcal{V}_i\oplus\mathcal{T}_i
$

\For{MoE layer $l{=}1,\ldots,L$}

\Statex
\textcolor{midgray}{\textit{// Step 2: Layer-wise thresholding}}

\State
    $\mathcal{S}{=}\{k|r_{k}^l{\geq}\tau_v^l~\text{for \textcolor{softblue}{[\textit{vis}]},}~\text{or}~r_{k}^l{\geq}\tau_t^l~\text{for \textcolor{orange}{[\textit{text}]}}\}$
\State
    $\hat{\mathbf y}_{ij}^l {=}\sum_{k{\in}\mathcal{S}} (r_k^l  \cdot e_k(\mathbf{\hat{x}}_{ij}^l))$ 
    \Comment{$k$: expert index}

\State    
    $\mathbf{y}_{ij}^l {=}\sum_{k=1}^{K} (r_k^l \cdot e_k(\mathbf{x}_{ij}^l))$ \Comment{reference}

\Statex
\textcolor{midgray}{\textit{// Step 3: Semantic-guided optimization}}
\State
    $L = \frac{1}{\sum_{i=1}^{N_s}|\mathcal{M}_i|}{
\sum_{i=1}^{N_s}
\|\mathcal{M}_i \odot (\mathcal{Y}_i^l-\hat{\mathcal{Y}}_i^l)\|_2^2}$

\State
    $
    (\tau_v^l,\tau_t^l)
    {=}
    \arg\min_{\tau_v^l,\tau_t^l}L
    $,~\textit{s.t.}~$\rho'{\ge} \rho$

\EndFor

\State \Return $
\{
(\tau_v^l,\tau_t^l)
\}_{l=1}^{L}
$
\end{algorithmic}
\end{algorithm}

\subsection{Semantic-guided Optimization}
\label{subsec:method_opt}
To monitor semantic drift of intermediate layers, \cc~calculates reconstruction error between default reference $\mathcal{Y}_i^l{=}\{\mathbf{y}_{ij}^l\}$ and skipped outputs $\mathcal{\hat{Y}}_i^l{=}\{\mathbf{\hat{y}}_{ij}^l\}$ within semantic mask. Let $ N_\mathrm {s}$ denote the number of samples in the calibration set. Given thresholds $(\tau_v^l,\tau_t^l)$ at layer $l$, we propose masked Mean Squared Error (MSE) to measure token-wise reconstruction error:
\begin{align}
L &= \frac{
\sum_{i=1}^{N_s}
\|\mathcal{M}_i \odot (\mathcal{Y}_i^l-\hat{\mathcal{Y}}_i^l)\|_2^2}{\sum_{i=1}^{N_s}|\mathcal{M}_i|},
\label{eq:mask_mse}
\end{align}
where the element-wise product $\odot$ restricts the MSE range within $\mathcal{M}_i$ (obtained from Eqn.~(\ref{eq:relevance_mask})), preventing noisy signals in the calibration.
The denominator $\sum_{i=1}^{N_{\mathrm{s}}}|\mathcal{M}_{i}|$ counts valid tokens after masking.
Given a target skipping ratio $\rho$, our objective is formulated as minimizing the masked MSE, subject to the achieved skipping ratio $\rho'$ under $(\tau_v^l,\tau_t^l)$ meeting or exceeding the desired $\rho$:
\begin{align}
    (\tau_v^l,\tau_t^l)
    &=
    \arg\min_{\tau_v^l,\tau_t^l}L
    ,~\textit{s.t.}~ \rho'\ge \rho.
\label{eq:obj}
\end{align}
Here, the skipping ratio is defined as the fraction \textit{w.r.t.} the originally activated Top-K experts, not all candidate experts.
When calibrating layer $l$, preceding layers apply their calibrated thresholds to account for error accumulation from upstream experts.
Compared with supervision from the end output \citep{huang2026modes}, \cc~probes the representation divergence of each layer under semantic guidance, achieving fine-grained and flexible parameterization. 
Notably, ``probe" generally refers to linear probes, where a simple linear classifier is trained on a frozen model to diagnose intermediate representations. 
Here, we use the broader sense of diagnosis, \textit{i.e.}, assessing the distortion of intermediate features through reconstruction error, while not training auxiliary modules (\textit{e.g.}, MLPs).

Our calibration procedure is summarized in Alg.~\ref{alg:ours}.
We construct semantic mask $\mathcal{M}_i$ for each sample $i$ in the projection space, then conduct skipping in each MoE layer. 
To distinguish from the default input $\mathbf{x}_{ij}^l$ ($j$ as token index), we denote $\hat{\mathbf{x}}_{ij}^l$ as the skipped input applying calibrated thresholds in preceding layers.
The reference branch converts token $\mathbf{x}_{ij}^l$ into $\mathbf{y}_{ij}^l$ with $K$ experts, while the skipped branch transforms $\mathbf{\hat{x}}_{ij}^l$ to $\mathbf{\hat{y}}_{ij}^l$ with selected experts.
We contrast reference and expert-skipped output by masked MSE and obtain $(\tau_v^l,\tau_t^l)$ in a grid search. 
The semantic guidance of our method is twofold: (i) constructing semantic masks to focus the optimization on salient tokens ($\mathcal{M}_i$), (ii) probing the semantic consistency of intermediate features 
($\mathcal{Y}_i^l, \mathcal{\hat{Y}}_i^l$).

\begin{table*}[ht]
\centering
\caption{\textbf{Performance on Qwen3VL-MoE-30B-A3B \citep{Qwen3VL}.} We report average scores (Avg) and relative percentage (Rel) to the default model, with actual skipping ratios (\note{${\rho'}$}) in subscripts.
ProSE achieves SOTA performance and skips more experts in practice.}
\label{tab:img_qwen3}
\resizebox{0.99\textwidth}{!}{
\begin{tabular}{lccccccccc|cc}
\toprule
Method & MMMU & SQA &  TextVQA & ChartQA & OCRBench & MMStar & RealWorldQA & VizWiz & InfoVQA  & Avg & 
Rel\note{$_{\rho'}$}(\%)  \\
\midrule
$K{=}8$ & 53.4 & 94.9 & 83.4 & 85.1 & 82.9 & 58.7 & 65.1 & 71.9 & 81.3  & 75.2 & 100.0 \\

\hline
\rowcolor{lightgray}
\multicolumn{12}{c}{\textit{skip $\rho{=}$63\%}} \\
\rowcolor{white}
${K{=}3}_\mathrm{top}$ & 51.7 & 93.2 & 80.8 & 78.0 & 84.3 & 61.2 & 46.4 & 69.7 & 76.1 & 71.3 & 94.8 \\
${K{=}3}_\mathrm{rand}$ & 26.2 & 14.0 & 23.6 & 19.9 & 26.2 & 14.8 & 20.0 & 16.6 & 20.4  & 20.2 & 26.9 \\
NAEE \citep{lu2024not} \ver{ACL'24} & 50.7 & 93.7 & 81.2 & 79.6 & 83.0 & 57.9  & 50.3 & 70.4 & 76.9 & 71.5 & 95.1 \\
DiEP \citep{bai2025diep} \ver{NeurIPS'25} & 50.1 & 93.9 & 81.8 & 80.1 & 82.7 & 57.5  & 55.2 & 71.1 & 77.2  & 72.2 & 96.0 \\
MoDES \citep{huang2026modes} \ver{CVPR'26} & 50.9 & \textbf{95.4} & 81.8 & 82.9 & 81.3 & \textbf{58.6} & 64.8 & \textbf{72.1} & 78.0 & 74.0 & 98.4\note{$_{62.9}$} \\
\rowcolor{mint}
\cc & \textbf{51.8} & 95.2 & \textbf{82.9} & \textbf{84.5} & \textbf{83.2} & 58.5 & \textbf{66.1} & 71.7 & \textbf{78.8} & \textbf{74.7} & \textbf{99.4}\note{$_{64.8}$} \\
\hline
\rowcolor{lightgray}
\multicolumn{12}{c}{\textit{skip $\rho{=}$75\%}} \\
\rowcolor{white}
${K{=}2}_\mathrm{top}$ & 50.3 & 89.7 & 77.1 & 70.7 & 81.5  & 58.0 & 54.8  & 66.5 & 72.8  & 69.0 & 91.8 \\
${K{=}2}_\mathrm{rand}$  & 25.3 & 13.4 & 20.5 & 17.1 & 20.5 & 12.4 & 17.3 & 15.9 & 17.7  & 17.8 & 23.7 \\
NAEE \citep{lu2024not} \ver{ACL'24} & 50.1 & 88.4 & 79.9 & 74.4 & 77.9 & 61.8 & 52.0 & 68.3 & 77.0 & 70.0 & 93.1 \\
DiEP \citep{bai2025diep} \ver{NeurIPS'25} & 51.2 & 88.9 & 80.4 & 74.9 & 78.4 & 62.7 & 52.7 & 68.8 & 77.9 & 70.7 & 94.0 \\
MoDES \citep{huang2026modes} \ver{CVPR'26} & 53.4 & \textbf{95.0 }& 81.4 & 81.0 & 80.5 & 57.6 & 66.0 & 71.6 & 75.6 & 73.6 & 97.8\note{$_{76.2}$} \\
\rowcolor{mint}
\cc & \textbf{55.8} & 94.6 & \textbf{83.1} & \textbf{84.8} & \textbf{81.6} & \textbf{65.9} & \textbf{66.5} & \textbf{71.8} & \textbf{78.1} & \textbf{75.8} & \textbf{100.8}\note{$_{76.2}$} \\

\hline
\rowcolor{lightgray}
\multicolumn{12}{c}{\textit{skip $\rho{=}$88\%}} \\
\rowcolor{white}
${K{=}1}_\mathrm{top}$ & 39.2 & 63.6 & 61.8 & 57.6 & 62.3 & 31.6 & 52.4 & 51.1 & 56.3 & 52.9 & 70.3 \\
${K{=}1}_\mathrm{rand}$ & 22.5 & 11.6 & 18.8 & 15.3 & 18.9 & 11.0 & 15.8 & 14.7 & 16.1 & 16.1 & 21.4 \\

NAEE \citep{lu2024not} \ver{ACL'24} & 49.4 & 87.2 & 75.4 &  63.2 & 79.5 & 56.6 & 54.4  & 62.0 &  67.8 & 66.2 & 88.0 \\
DiEP \citep{bai2025diep} \ver{NeurIPS'25} & 50.9 & 88.7 & 76.1 &  64.8 & 79.8 & 56.9 & 59.2  & 65.7 &  67.8  & 67.8 & 90.2 \\
MoDES \citep{huang2026modes} \ver{CVPR'26} & 51.7 & \textbf{92.8} & 79.4 & 76.0 & 75.6 & 55.8 & 63.7 & \textbf{70.8}  & 69.3  & 70.6 & 93.9\note{$_{86.1}$} \\
\rowcolor{mint}
\cc  & \textbf{53.0} & 92.4 & \textbf{80.7} & \textbf{78.9} & \textbf{82.1} & \textbf{58.6} & \textbf{63.8} & {69.7} & \textbf{75.0} & \textbf{72.7} & \textbf{96.7}\note{$_{88.3}$} \\

\hline
\rowcolor{lightgray}
\multicolumn{12}{c}{\textit{skip $\rho{=}$95\%}} \\
\rowcolor{white}
NAEE \citep{lu2024not} \ver{ACL'24} & 30.1 & 55.9 & 47.4 & 41.6 & 42.8 & 21.6  & 46.5 & 15.5 & 28.6 & 36.7 & 48.8 \\
DiEP \citep{bai2025diep} \ver{NeurIPS'25} & 31.4 & 58.0 & 49.2 & 41.9 & 44.1 & 23.0  & 49.2 & 18.1 & 29.8  & 38.3 & 50.9 \\
MoDES \citep{huang2026modes} \ver{CVPR'26} & 30.3 & 42.5 & 13.6 & 0.9 & 2.2 & \textbf{29.8} & 6.0 & 42.1 & 14.6 & 20.2 & 26.9\note{$_{92.8}$} \\
\rowcolor{mint}
\cc & \textbf{36.7} & \textbf{72.5} & \textbf{52.0} & \textbf{46.0} & \textbf{65.5} & 23.4 & \textbf{51.9} & \textbf{48.5} & \textbf{45.1} & \textbf{49.1} & \textbf{65.3}\note{$_{96.0}$} \\

\bottomrule
\end{tabular}
}
\end{table*}

\paragraph{Implementation.} 
For $\tau_{v,t}^l {\in} (0,1)$, the default grid search over all candidate pairs corresponds to a 2D grid.
To reduce the number of candidate pairs, MoDES \citep{huang2026modes} assumes monotonicity of the objective with respect to thresholds, yielding a 1D frontier search. The rationale is that higher thresholds skip more experts, thereby leading to larger divergence with respect to the no-skipping reference. In this way, the frontier search only binds $\min \tau_v$ satisfying $\rho'{\geq}\rho$ to $\tau_t$, resulting in an efficient 1D search in the frontier area. Following this, we compute the masked MSE (Eqn.~(\ref{eq:mask_mse})) from the reduced candidate set, where $\tau_t^l$ only associates with min $\tau_v^l$ satisfying skipping ratios.

\subsection{Inference}
\label{subsec:method_infer}

During inference, \cc~applies calibrated thresholds and only activates necessary experts as described in Eqn.~(\ref{eq:rout_ours}). Specifically, at layer $l$, it skips experts with routing weights below $\tau_v^l$ for visual tokens and $\tau_t^l$ for text tokens.
Without extra modulation \citep{huang2026modes} or proxy estimation \citep{cai2026gemoe}, our approach achieves efficient and effective skipping in inference. Meanwhile, \cc~shows better compatibility with the token pruning techniques. See Sec.~\ref{subsec:exp_result} for the quantitative results. 
\section{Experiments}
\label{sec:exp}

\subsection{Setups}
\label{subsec:exp_setup}

\paragraph{Models and baselines.} 
We choose two MoE-MLLMs to conduct expert skipping:  
(i) Qwen3VL-MoE-30B-A3B-Instruct \citep{Qwen3VL} activates 8 from 128 experts in 48 MoE layers; (ii) KimiVL-A3B-Instruct \citep{kimiteam2025kimivl} has 26 MoE layers, each activating 6 from 64 routing experts and 2 shared experts. 
We compare \cc~with (i) vanilla approaches: random and Top-K retention; (ii) calibration-based methods: MoDES \citep{huang2026modes} for MoE-MLLMs, NAEE \citep{lu2024not} and DiEP \citep{bai2025diep} for MoE-LLMs.
Regarding practical skipping ratios $\rho'$, we report average values for MoDES \citep{huang2026modes} and ours, as other baselines show little variance. 
We provide per-dataset results in the appendix.

\begin{table*}[ht]
\centering
\caption{\textbf{Performance on KimiVL-A3B-Inst \citep{kimiteam2025kimivl}.} We report results given the target skipping ratio $\rho{=}\{67\%, 83\%, 95\%\}$. Our performance gain becomes more pronounced under aggressive settings (\textit{e.g.}, 95\% skipping).}
\label{tab:img_kimi}
\resizebox{0.99\textwidth}{!}{
\begin{tabular}{lccccccccc|cc}
\toprule

Method & MMMU & SQA & TextVQA & ChartQA & OCRBench & MMStar & RealWorldQA & VizWiz & InfoVQA & Avg & 
Rel\note{$_{\rho'}$}(\%)  \\
\hline
$K{=}6$ & 51.7 & 96.2 & 88.7 & 89.5 & 84.3 & 49.4 & 65.8 & 70.4 & 84.0 & 75.6 & 100.0 \\
\hline
\rowcolor{lightgray}
\multicolumn{12}{c}{\textit{skip $\rho{=}$67\%}} \\
\rowcolor{white}
${K{=}2}_\mathrm{top}$  & 46.7 & 92.8 & 85.9 & 87.4 & 80.7 & 54.2 & 63.8 & 68.7 & 76.8 & 73.0 & 96.6 \\
${K{=}2}_\mathrm{rand}$  & 26.1 & 50.3 & 55.1 & 52.6 & 43.0 & 32.5 & 42.1 & 49.6 & 43.8 & 43.9 & 58.1 \\
NAEE \citep{lu2024not} \ver{ACL'24} & 46.1 & 92.9 &  82.8 & 84.9 & 81.4 & 53.8 & 60.5 & 69.1 & 76.2 & 72.0 & 95.2 \\
DiEP \citep{bai2025diep} \ver{NeurIPS'25} & 47.8 & 93.4 &  83.2 & 85.4 & 82.6 & \textbf{55.1} & 63.7 & \textbf{69.4} & 77.3 & 73.1 & 96.7 \\
MoDES \citep{huang2026modes} \ver{CVPR'26} & \textbf{52.1} & \textbf{95.7} & 85.4 & 87.9 & 81.1 & 48.8 & \textbf{64.7} & 69.3 & 79.4 & 73.8 & 97.7\note{$_{67.7}$} \\
\rowcolor{mint}
\cc & 50.3 & 95.0 & \textbf{87.3} & \textbf{88.8} & \textbf{82.7} & 51.7 & 63.5 & {69.3} & \textbf{80.8} & \textbf{74.4} & \textbf{98.4}\note{$_{68.0}$} \\

\hline
\rowcolor{lightgray}
\multicolumn{12}{c}{\textit{skip $\rho{=}$83\%}} \\
\rowcolor{white}
${K{=}1}_\mathrm{top}$ & 40.2 & 77.2 & 77.9 & 78.4 & 69.1 & 43.1 & 54.4 & 66.1 & 71.0 & 64.2 & 84.9 \\
${K{=}1}_\mathrm{rand}$ & 25.7 & 11.1 & 14.9 & 20.5 & 18.4 & 21.7 & 16.5 & 10.2 & 15.2 & 17.1 & 22.6 \\
NAEE \citep{lu2024not} \ver{ACL'24} & 38.9  & 78.1  & 77.1 & 78.2  & 67.5 & 40.2 & 54.9  & 66.0 & 72.8  & 63.7 & 84.3 \\
DiEP \citep{bai2025diep} \ver{NeurIPS'25} & 39.2  & 78.8  & 78.2 & 80.0  & 69.3 & 41.3 & 58.5  & 66.8 & 73.3  & 65.0 & 86.1 \\
MoDES \citep{huang2026modes} \ver{CVPR'26} & \textbf{50.4} & 92.7 & 81.6 & 84.4 & 75.6 & 46.7 & 64.4 & 67.2 & 73.6 & 70.8 & 93.6\note{$_{81.5}$} \\
\rowcolor{mint}
\cc & 49.2  & \textbf{93.4}  & \textbf{85.2} & \textbf{87.0}  & \textbf{79.4} & \textbf{50.7} & \textbf{64.7}  & \textbf{68.5}  & \textbf{74.7} & \textbf{72.5} & \textbf{96.0}\note{$_{84.5}$} \\
\hline
\rowcolor{lightgray}
\multicolumn{12}{c}{\textit{skip $\rho{=}$95\%}} \\
\rowcolor{white}
NAEE \citep{lu2024not} \ver{ACL'24} & 27.6 & 13.2 & 0.8 & 0.0 & 12.5 & 2.9 & 0.8 & 18.9 & 0.2 & 8.5 & 11.2 \\
DiEP \citep{bai2025diep} \ver{NeurIPS'25} & 28.8 & 13.9 & 1.4 & 0.5 & 14.8 & 4.2 & 3.1 & 20.5 & 1.7 & 9.9 & 13.1 \\
MoDES \citep{huang2026modes} \ver{CVPR'26} & 41.9 & \textbf{84.8} & 47.0 & 49.8 & 27.6 & 39.2 & \textbf{60.5} & 49.7 & 40.7 & 49.0 & 64.8\note{$_{95.7}$}\\
\rowcolor{mint}
\cc & \textbf{42.3}  & 82.1 & \textbf{68.3} & \textbf{54.9} & \textbf{59.5} & \textbf{49.2} & 59.9 & \textbf{58.6} & \textbf{45.3} & \textbf{57.8} & \textbf{76.5}\note{$_{96.0}$} \\

\bottomrule
\end{tabular}
}
\end{table*}

\begin{table*}[h]
\centering
\caption{\textbf{Combination with visual token pruning.} 
ProSE is more compatible with token pruning when combined with MI-Pruner (\texttt{+MI}) \citep{li2026mi} and DivPrune (\texttt{+Div}) \citep{alvar2025divprune}, outperforming MoDES with average performance gains of $2\%{\sim}5\%$.
}
\label{tab:prune}
\resizebox{0.99\textwidth}{!}{
\begin{tabular}{lcccccc|lcccccc}
\hline
Method & TextVQA & ChartQA & OCRBench & MMStar & InfoVQA & Rel (\%) &
Method & TextVQA & ChartQA & OCRBench & MMStar & InfoVQA & Rel (\%) \\
\hline
\multicolumn{7}{c|}{\textit{Qwen3VL-MoE-30B-A3B}} &
\multicolumn{7}{c}{\textit{KimiVL-A3B-Inst}} \\
\hline
$K{=}8$ & 83.4 & 85.1 & 82.9 & 58.7 & 81.3 & 100.0 &
$K{=}6$ & 88.7 & 89.5 & 84.3 & 49.4 & 84.0 & 100.0 \\
\hline
\rowcolor{lightgray}
\multicolumn{7}{c|}{\textit{skip $\rho{=}$75\%, prune 20\%}} &
\multicolumn{7}{c}{\textit{skip $\rho{=}$67\%, prune 20\%}} \\
\rowcolor{white}
MoDES\texttt{+MI} & 81.6 & 80.1 & 77.8 & 56.9 & 74.9 & 94.9 & 
MoDES\texttt{+MI}  & 85.2 &  86.3 &  73.6  & 48.7  & 78.1 & 93.9 \\
\cc\texttt{+MI} & 82.8 & 83.8 & 79.8 & 66.2 & 77.3 & 99.6 & 
\cc\texttt{+MI} & 87.1  & 87.3 &  75.8  & 50.9  & 78.1 & 95.7 \\
MoDES\texttt{+Div} & 81.5 & 80.0 & 79.4 & 56.5 & 74.8 & 95.1 & 
MoDES\texttt{+Div} & 83.5 & 84.9 & 74.1 & 46.9 & 74.3 & 91.9 \\
\cc\texttt{+Div} & 82.7 & 83.9 & 81.6 & 65.4 & 77.2 & 99.8 & 
\cc\texttt{+Div} & 86.8 & 86.7 & 76.3 & 50.0 & 76.1 & 94.9 \\
\hline
\rowcolor{lightgray}
\multicolumn{7}{c|}{\textit{skip $\rho{=}$88\%, prune 20\%}} &
\multicolumn{7}{c}{\textit{skip $\rho{=}$83\%, prune 20\%}} \\
\rowcolor{white}
MoDES\texttt{+MI} & 79.5 & 74.6 & 73.7 & 55.3 & 68.2 & 89.8 & 
MoDES\texttt{+MI}  & 81.5 & 82.9 & 67.5 & 47.1 & 70.1 & 88.1 \\
\cc\texttt{+MI} & 80.9 & 77.8 & 79.7 & 58.0 & 73.3 & 94.5 & 
\cc\texttt{+MI} & 84.5 & 85.5 & 72.2 & 50.3 & 71.7 & 91.9 \\
MoDES\texttt{+Div} & 79.6 & 75.2 & 75.9 & 56.5 & 67.7 & 90.7 & 
MoDES\texttt{+Div} & 81.3 & 81.2 & 69.3 & 46.8 & 64.8 & 86.7 \\
\cc\texttt{+Div} & 80.9 & 78.4 & 81.6 & 57.7 & 73.8 & 95.1 & 
\cc\texttt{+Div} & 84.3 & 84.8 & 73.9 & 49.1 & 70.4 & 91.6 \\
\bottomrule
\end{tabular}
}
\end{table*}

\paragraph{Datasets.} 
For calibration, we randomly collect 128 samples from GQA~\citep{hudson2019gqa} (train-set).
After calibration, we evaluate expert skipping approaches on nine visual question answering (VQA) benchmarks: (i) knowledge-based reasoning: MMMU \citep{yue2023mmmu}, SQA-IMG \citep{lu2022learn} (abbreviated as SQA in tables); (ii) general visual understanding: VizWiz \citep{gurari2018vizwiz}, MMStar \citep{chen2024we} and RealWorldQA \citep{ai2024grok}; (iii) textual and structural recognition: TextVQA \citep{singh2019towards}, ChartQA \citep{masry2022chartqa}, OCRBench \citep{liu2024ocrbench} and InfoVQA \citep{mathew2022infovqa}.
The evaluation follows the standard settings of LMMs-Eval \citep{zhang2025lmms} using Flash-Attention2 \citep{dao2023flashattention2}.

\paragraph{Implementation details.}
We run experiments on 4 A100 GPUs and perform a grid search with 50 points in $(0,1)$. 
We reproduce MoDES \citep{huang2026modes} from their official repository and adopt the skipping rules of NAEE \citep{lu2024not} and DiEP \citep{bai2025diep} for MoE-MLLMs. All methods are calibrated with the same set of 128 GQA samples.
Notably, the original MoDES paper sets $\rho$ slightly below the reported skipping ratio, \textit{e.g.}, set $\rho{=}65\%$ for 67\% skipping. To avoid additional tuning, we simply set $\rho$ equal to the desired skipping ratio consistently, \textit{e.g.}, $\rho{=}67\%$ for 67\% skipping.

\subsection{Results}
\label{subsec:exp_result}

\paragraph{Qwen3VL-MoE-30B-A3B.}
Inspired by random and salience-based token pruning \citep{li2026mi,wen2025stop}, we include random and Top-K expert retention as two vanilla baselines. 
As shown in Tab.~\ref{tab:img_qwen3}, ${K{=}3}_\mathrm{top}$ retains 3 experts with highest routing weights, and ${K{=}3}_\mathrm{rand}$ randomly selects 3 experts from 8 experts in Qwen3VL-MoE-30B-A3B.
Although random pruning shows comparable performance \citep{li2026mi,wen2025stop}, random skipping causes severe degradation. This indicates that experts have more specialized roles, which is different from the subset coverage problem in token pruning. As a result, inappropriate routing and skipping undermine performance substantially. Despite the calibration process in NAEE \citep{lu2024not} and DiEP \citep{bai2025diep}, their performance becomes worse as $\rho$ increases.
In contrast, \cc~achieves averagely 100.8\% performance of the default model when $\rho{=}75\%$ and maintains 65.3\% performance under the aggressive skipping of $\rho{=}95\%$.
Compared with MoDES, our layer-wise calibration is more flexible and expressive
to reflect the individual sensitivity of experts. By probing reconstruction error under a constraint $\rho$ at each layer, our method ensures SOTA performance and a compliant skipping ratio in practice, which is always higher than MoDES. Please refer to 
the appendix for more discussion.

\paragraph{KimiVL-A3B-Inst.} For KimiVL-A3B-Inst \citep{kimiteam2025kimivl}, we skip from $K{=}6$ routing experts in 26 MoE layers while keeping two shared experts as they are. In Tab.~\ref{tab:img_kimi}, \cc~outperforms MoDES and other baselines by a substantial margin. The advantage becomes more pronounced as the skipping ratio increases. Specifically, \cc~maintains 76.5\% performance under 95\% skipping, which is higher than the same skipping ratio on Qwen3VL-MoE-30B-A3B (65.3\%). This resilience might be attributed to the shared experts in KimiVL-A3B-Inst, which provide a fallback when most routing experts are skipped.
Although NAEE \citep{lu2024not} and DiEP \citep{bai2025diep} also use layer-wise thresholds, their performance is much lower than ours, and not necessarily better than the Top-K skipping in Tab.~\ref{tab:img_kimi}.
The performance gap stems from the calibration objective.
Their calibration \citep{lu2024not,bai2025diep} focuses solely on achieving the target skipping ratio, whereas ours aims to minimize layer-wise semantic drift through reconstruction error before and after skipping.

\paragraph{Combination with token pruning.} 
Expert skipping can be combined with token pruning techniques for further speedup. Here, we prune 20\% visual tokens using MI-Pruner \citep{li2026mi} and DivPrune \citep{alvar2025divprune} in the projection space. To show the robustness, we apply the same thresholds as in previous sections without recalibration under token pruning.
As reported in Tab.~\ref{tab:prune}, \cc~outperforms MoDES consistently across different settings. With more than 80\% skipping and 80\% token retention, our method maintains ${\geq}$94.5\% performance on Qwen3VL and ${\geq}$91.6\% on KimiVL models.
The benefits stem from our explicit incorporation of visual importance during calibration (Sec.~\ref{subsec:method_token}).
Concretely, token pruning selects semantically important tokens as LLM inputs, and our calibration ensures minimum reconstruction error of these important tokens after expert skipping, showing the same spirit. 
We provide additional results under more pruning ratios in 
the appendix.

\paragraph{Efficiency.} 
In Tab.~\ref{tab:eff}, we report calibration cost, average prefill time and Time-To-First-Token (TTFT) using the first 200 samples from ChartQA \citep{masry2022chartqa} with batch size${=}$1. The last column shows the performance score and actual skipping ratios (${\rho'}$) of ChartQA. The default KimiVL-A3B-Inst without skipping is denoted as ``w/o".
MoDES employs a two-stage calibration strategy to estimate layer importance (2.5 min) and thresholds (4.9 min), taking 7.4 min in total. Although \cc~takes longer (5.1 min) to calibrate layer-wise thresholds, its overall time is shorter than two-stage MoDES. At inference time, our method achieves a higher actual expert skipping ratio, resulting in shorter prefill latency and TTFT.

\begin{table}[h]
\centering
\caption{\textbf{Efficiency comparison.} We accelerate the prefill and TTFT by $1.7{\times}$ and $1.4{\times}$ on KimiVL-A3B-Inst ($\rho{=}83\%$).}
\label{tab:eff}
\resizebox{0.48\textwidth}{!}{
\begin{tabular}{lcccc}
\hline
Method  & Calib. (min) & Prefill (ms) & TTFT (ms)  & Perform$_{\rho'}$ \\
\hline
w/o & - & 261.8 & 293.5 & 89.5$_{0\%}$ \\
MoDES & 2.5+4.9 & 155.6 & 219.3  & 84.4$_{84.2\%}$ \\
\rowcolor{mint}
\cc & 5.1 &  149.7 \textbf{($1.7\times$)} & 208.3 \textbf{($1.4\times$)} & 87.0$_{85.6\%}$ \\
\bottomrule
\end{tabular}
}
\end{table}

\subsection{Ablation Study}
\label{subsec:exp_ablation}

\paragraph{Calibration data.} We adopt 128 samples from GQA \citep{hudson2019gqa} as the calibration set. In Tab.~\ref{tab:ablation_coco_token} (top), we compare with 128 samples from COCO-Caption 2017 \citep{chen2015microsoft,lin2014microsoft} (validation set), and \{64, 512\} samples from GQA \citep{hudson2019gqa}. As observed, using COCO$_\mathrm{128}$ as the calibration set leads to a slight drop in overall performance, which may be attributed to the distribution gap between captioning and VQA data. Moreover, increasing the calibration set size (GQA$_\mathrm{512}$) yields only limited improvement, since our layer-wise calibration decomposes the global thresholding into per-layer problems that are easy to solve with few samples.

\begin{table}[!h]
\centering
\caption{\textbf{Ablation study on calibration data and importance metrics.} Reported on Qwen3VL-MoE-30B-A3B with $\rho{=}88\%$.}
\label{tab:ablation_coco_token}
\resizebox{0.49\textwidth}{!}{
\begin{tabular}{lcccccc}
\hline
Method  & MMMU & SQA  & TextVQA & ChartQA & MMStar & Rel (\%) \\
\hline
$K{=}8$ & 53.4 & 94.9 & 83.4 & 85.1 & 58.7 & 100.0 \\
\hline
COCO$_\mathrm{128}$ & 51.0  & 90.4  &  80.0  & 76.6  & 57.5  & 94.7 \\
GQA$_\mathrm{64}$ & 51.5  & 91.5  &  80.4  & 78.8  & 57.8  & 95.9 \\
GQA$_\mathrm{512}$ & 52.9 & 92.8 & 80.9 & 79.0  & 58.6 & 97.0 \\
\hline
Uniform & 49.6 & 91.5 & 79.5 &  76.9 & 57.7 & 94.6 \\
Random & 51.3 & 92.1 & 80.2 &  78.1 & 58.2 &  95.8 \\
VisPruner \citep{zhang2025beyond} & 50.8 & 89.9 & 80.0 & 75.9 & 58.0 & 94.4 \\
DivPrune \citep{alvar2025divprune} & 52.4 & 91.6 & 80.4 & 78.5 & 58.5 &  96.2\\
\rowcolor{mint}
Ours & 53.0 & 92.4 & 80.7 & 78.9  & 58.6 & 96.8 \\
\bottomrule
\end{tabular}
}
\end{table}

\begin{figure}[h]
\centering
\begin{subfigure}[h]{0.44\linewidth}
    \centering
    \includegraphics[width=\linewidth]{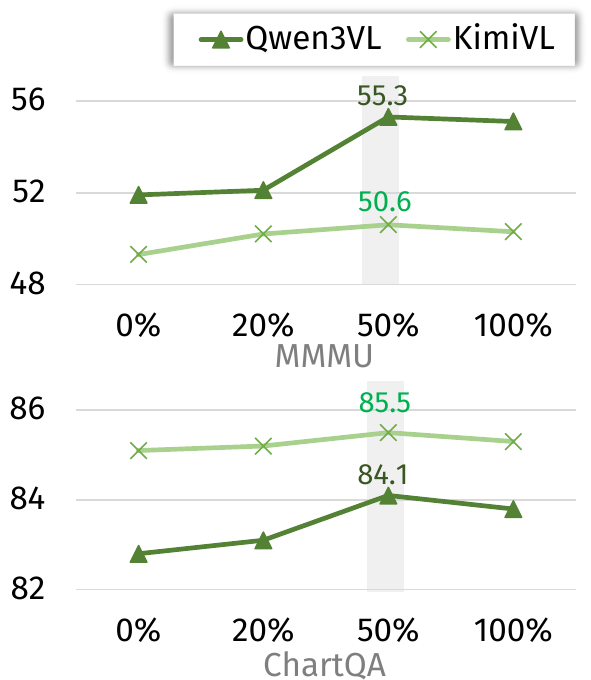}
    \caption{Skip 70\% experts}
    \label{fig:beta_70}
\end{subfigure}
\centering
\begin{subfigure}[h]{0.44\linewidth}
    \centering
    \includegraphics[width=\linewidth]{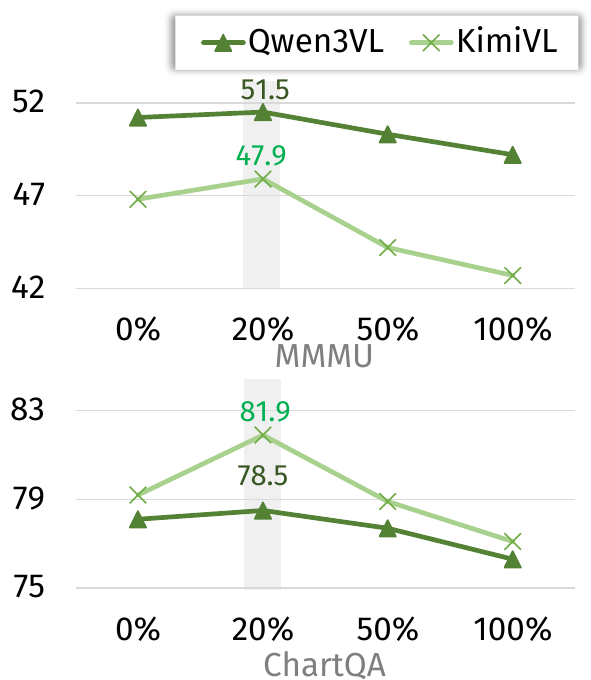}
    \caption{Skip 90\% experts}
    \label{fig:beta_90}
\end{subfigure}
\vspace{-0.3cm}
\caption{\textbf{Ablation study on $\beta_v$.} We keep $\beta_v{=}50\%$ for general skipping (a), and $20\%$ for aggressive cases (b).
}
\label{fig:ablation_beta}
\vspace{-0.2cm}
\end{figure}

\vspace{-0.3cm}
\paragraph{Visual ratios $\beta_v$.} 
For semantic masks, we retain 50\% visual tokens for general skipping when $\rho{\leq}85\%$ and 20\% for aggressive cases when $\rho{>}85\%$. The rationale is that aggressive skipping causes severe degradation, making reconstruction error noisy and requiring a more semantically concentrated mask. 
We enumerate $\beta_v{=}\{0\%, 20\%, 50\%, 100\%\}$ on MMMU \citep{yue2023mmmu} and ChartQA \citep{masry2022chartqa} in Fig.~\ref{fig:ablation_beta}, where 0\% means no visual tokens attend calibration objectives, and 100\% means all visual tokens without masking. 
As observed, $\beta_v{=}50\%$ shows optimal results under 70\% skipping in (a) and $\beta_v{=}20\%$ is the best choice under 90\% skipping in (b).
As described in Sec.~\ref{subsec:method_token}, the mask's first effect is \textit{visual down-weighting}, and $\beta_v$ acts as a \textit{dial} that steers calibration towards preserving consistent vision understanding or language capacity.

\paragraph{Token importance metrics.}
Regarding the second effect of \textit{prioritizing important tokens}, we compare with uniform, random and two pre-LLM token scoring methods: (i) VisPruner \citep{zhang2025beyond} with attention salience; (ii) DivPrune \citep{alvar2025divprune} with Max-Min diversity. Here, ``uniform" downweights the visual MSE uniformly with the same $\beta_v$, replacing masked MSE with modality-weighted MSE.
As shown in Tab.~\ref{tab:ablation_coco_token}, random is a comparable baseline for overall semantic coverage \citep{wang2026token,wen2025stop}, 
while uniform degrades more by acting merely as a modality dial, validating the necessity of our semantic mask.
The attention-based VisPruner is less reliable due to positional bias and sink tokens \citep{wen2025stop,zhang2025cdpruner}, and we provide more discussion in 
the appendix.
DivPrune performs closest to ours, while our simple Top-K is more efficient than its iterative greedy selection.
Nevertheless, these scoring methods work only in the calibration process, and differ from ours by at most 2.4\%, showing less sensitivity than test-time token pruning evaluation \citep{alvar2025divprune,zhang2025beyond}.

\paragraph{Progressive search.}
We calibrate the skipping thresholds progressively, \textit{i.e.}, when searching for thresholds at layer $l$, early-layer experts (${<}l$) are skipped by searched thresholds ($\tau_{v,t}^{<l}$). In this way, the error accumulation is integrated into the calibration process.
In Tab.~\ref{tab:ablation_acc}, we report the results of (i) ``w/o early'': search for $\tau_{v,t}^l$ without early expert skipping; (ii) ``w/o layer": search for global $\tau_{v,t}$ by the final-layer masked MSE, without layer-wise calibration. As observed, ignoring $\tau_{v,t}^{<l}$ shows a slight degradation within 1\%, while totally omitting layer-wise thresholding leads to a substantial drop up to 9.3\% (skip 83\%).

\begin{table}[t]
\centering
\caption{\textbf{Ablation study on progressive search (KimiVL-A3B-Inst).} ``w/o early": without applying thresholds from preceding layers; ``w/o layer": without layer-wise calibration.
}
\label{tab:ablation_acc}
\resizebox{0.48\textwidth}{!}{
\begin{tabular}{lcccccc}
\hline
Method  & MMMU & SQA  & TextVQA & ChartQA & MMStar & Rel (\%) \\
\hline
\rowcolor{lightgray}
\multicolumn{7}{c}{\textit{skip $\rho{=}$67\%}} \\
\rowcolor{white}
w/o early & 49.9 & 95.1 & 86.6 & 88.0  & 51.1 & 99.4 \\
w/o layer & 43.6 & 87.5 & 82.0 & 82.7  & 47.3  & 93.8 \\
\rowcolor{mint}
Ours & 50.3 & 95.0 & 87.3 & 88.8  & 51.7 & 100.0 \\
\hline
\rowcolor{lightgray}
\multicolumn{7}{c}{\textit{skip $\rho{=}$83\%}} \\
\rowcolor{white}
w/o early & 48.7 & 92.6 & 84.9 & 86.6 & 49.7 & 99.2 \\
w/o layer & 42.1 & 86.4 & 80.5 & 80.2 & 42.4  & 90.7 \\
\rowcolor{mint}
Ours & 49.2 & 93.4 & 85.2 & 87.0  & 50.7 & 100.0 \\
\bottomrule
\end{tabular}
}
\end{table}

\subsection{Further Study}
\label{subsec:exp_further}

\begin{figure}[h]
\centering
    \includegraphics[width=\linewidth]{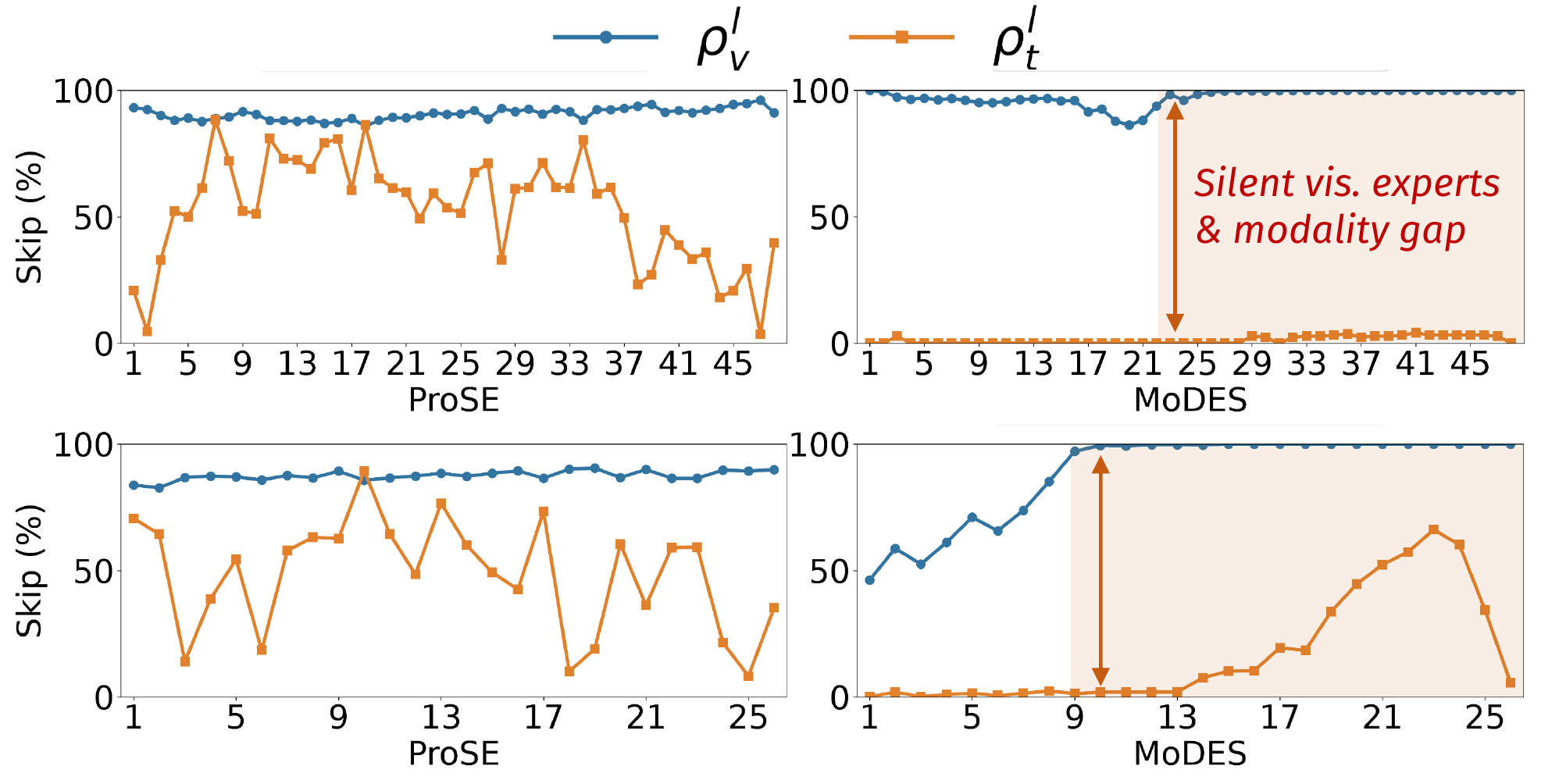}
\caption{
\textbf{Practical skipping ratios} for visual tokens ($\rho_v^l$) and textual tokens ($\rho_t^l$). 
ProSE shows dynamic skipping, while MoDES keeps visual experts silent from middle layers. 
\textbf{Top:} Qwen3VL (${\rho{=}0.88}$), \textbf{Bottom:} KimiVL (${\rho{=}0.83}$).
}
\label{fig:rho}
\end{figure}

\paragraph{Skipping ratios in practice.}
We provide the modality-specific skipping ratios $\rho_{v,t}^l$ across layers in Fig.\ref{fig:rho} (left: ProSE, right: MoDES).
The skipping ratio is tested on the first 200 samples in ChartQA \citep{masry2022chartqa}. Both \cc~and MoDES skip more experts for visual tokens (always blue on the top), while MoDES retains very few visual experts since the middle layers, creating pronounced modality imbalance. On Qwen3VL-MoE-30B-A3B (top, right), MoDES mutes experts serving visual tokens (skip ${\sim}95\%$) from layer 23 on, but textual tokens retain most of their experts (skip ${\sim}5\%$). 
The dormant visual experts in MoDES may undermine visual perception and reasoning.
In comparison, \cc~yields a relatively stable visual skipping ratio across layers and dynamic textual ratios from the layer-wise calibration. See 
our appendix for more discussion.

\begin{figure}[h]
\centering
\begin{subfigure}[h]{\linewidth}
    \centering
    \includegraphics[width=\linewidth]{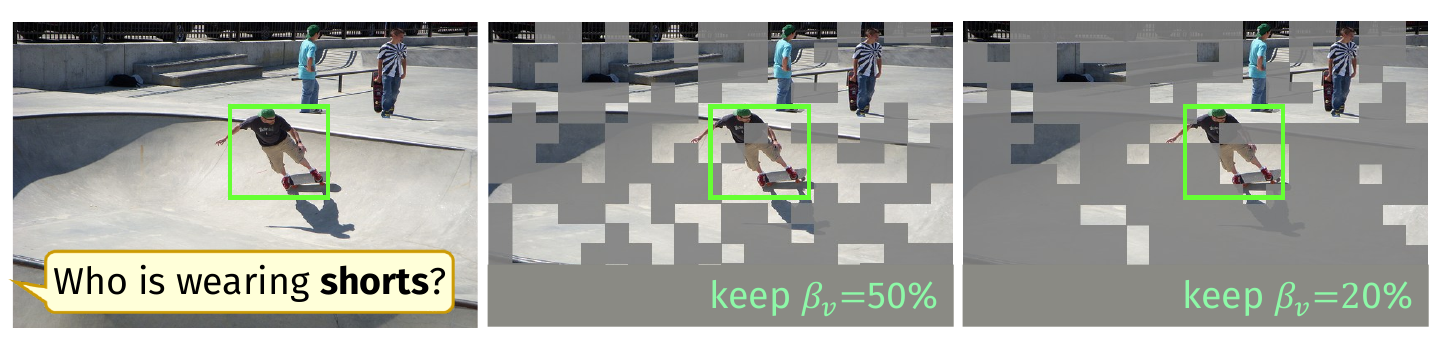}
    \caption{Qwen3VL-MoE-30B-A3B}
    \label{fig:mask_c}
\end{subfigure}
\begin{subfigure}[h]{\linewidth}
    \centering
    \includegraphics[width=\linewidth]{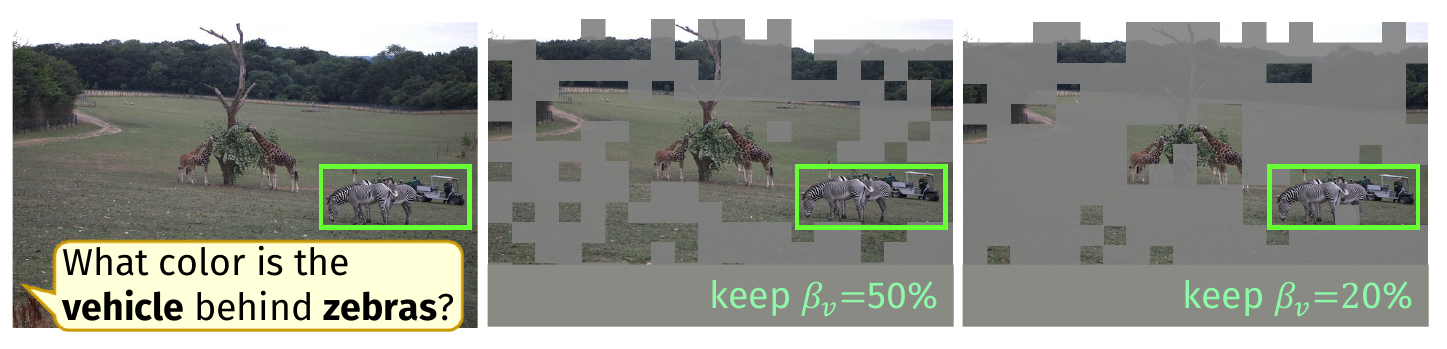}
    \caption{KimiVL-A3B-Inst}
    \label{fig:mask_d}
\end{subfigure}
\vspace{-0.3cm}
\caption{\textbf{Token masks} constructed from GQA \citep{hudson2019gqa}. The mask provides semantic guidance for layer-wise optimization.
}
\label{fig:mask_exp}
\vspace{-0.3cm}
\end{figure}

\paragraph{Visualization.}
Fig.~\ref{fig:mask_exp} visualizes retained and excluded tokens. The max aggregation over text tokens in Eqn.~(\ref{eq:relevance_cross}) points to the most relevant words (in bold) in the query and provides strong semantic guidance.
By retaining semantically relevant tokens while filtering out redundant ones, we keep the calibration objective focused on the consistency of task-relevant positions. 
Acting as a dial, the smaller ratio $\beta_v{=}20\%$ further enforces visual consistency on important patches with less background.
Moreover, our semantic-guided calibration is highly compatible with token pruning methods \citep{li2026mi} in Tab.~\ref{tab:prune}, further improving efficiency.

\section{Conclusion}
\label{sec:conclusion}

We propose \cc, a semantic-guided expert skipping method for MoE-MLLMs. Our approach stems from two observations: (i) visual tokens exhibit lower sensitivity than the text counterparts; (ii) the skipping-induced error accumulates across layers. Built upon these findings, we probe layer-wise reconstruction error and minimize the semantic drift caused by skipping to balance the performance \textit{vs.} efficiency trade-off.
Extensive experiments on 9 benchmarks and two MoE-MLLMs validate our effectiveness.
Empirically, \cc~preserves 96\% performance under 83\% expert skipping and accelerates the prefill time by $1.7{\times}$ on KimiVL-A3B-Inst.

\paragraph{Limitations and future work.} 
Although compatible with token pruning methods, retained tokens still pass through all layers with associated overhead. Reducing the depth of computation, such as early exit \citep{tan2026decomoe}, could further improve inference efficiency.



{
    \small
    \bibliographystyle{ieeenat_fullname}
    \bibliography{main}

@String(CVPR= {IEEE Conf. Comput. Vis. Pattern Recog.})

@String(ICCV= {Int. Conf. Comput. Vis.})

@String(ECCV= {Eur. Conf. Comput. Vis.})

@String(ICLR = {Int. Conf. Learn. Represent.})

@String(CVPR  = {CVPR})

@String(ICCV  = {ICCV})

@String(ECCV  = {ECCV})

@String(ICLR  = {ICLR})

@InProceedings{huang2026modes,
    title = {MoDES: Accelerating Mixture-of-Experts Multimodal Large Language Models via Dynamic Expert Skipping}, 
    author = {Yushi Huang and Zining Wang and Zhihang Yuan and Yifu Ding and Ruihao Gong and Jinyang Guo and Xianglong Liu and Jun Zhang},
    booktitle = {CVPR},
    year = {2026},
}

@inproceedings{ji2025specvlm,
  title={Specvlm: Enhancing speculative decoding of video llms via verifier-guided token pruning},
  author={Ji, Yicheng and Zhang, Jun and Xia, Heming and Chen, Jinpeng and Shou, Lidan and Chen, Gang and Li, Huan},
  booktitle={EMNLP},
  year={2025}
}

@article{jiang2024mixtral,
  title={Mixtral of experts},
  author={Jiang, Albert Q and Sablayrolles, Alexandre and Roux, Antoine and Mensch, Arthur and Savary, Blanche and Bamford, Chris and Chaplot, Devendra Singh and Casas, Diego de las and Hanna, Emma Bou and Bressand, Florian and others},
  journal={arXiv preprint arXiv:2401.04088},
  year={2024}
}

@article{wu2024deepseek,
  title={Deepseek-vl2: Mixture-of-experts vision-language models for advanced multimodal understanding},
  author={Wu, Zhiyu and Chen, Xiaokang and Pan, Zizheng and Liu, Xingchao and Liu, Wen and Dai, Damai and Gao, Huazuo and Ma, Yiyang and Wu, Chengyue and Wang, Bingxuan and others},
  journal={arXiv preprint arXiv:2412.10302},
  year={2024}
}

@article{liu2024deepseek,
  title={Deepseek-v2: A strong, economical, and efficient mixture-of-experts language model},
  author={Liu, Aixin and Feng, Bei and Wang, Bin and Wang, Bingxuan and Liu, Bo and Zhao, Chenggang and Dengr, Chengqi and Ruan, Chong and Dai, Damai and Guo, Daya and others},
  journal={arXiv preprint arXiv:2405.04434},
  year={2024}
}

@article{yang2025qwen3,
  title={Qwen3 technical report},
  author={Yang, An and Li, Anfeng and Yang, Baosong and Zhang, Beichen and Hui, Binyuan and Zheng, Bo and Yu, Bowen and Gao, Chang and Huang, Chengen and Lv, Chenxu and others},
  journal={arXiv preprint arXiv:2505.09388},
  year={2025}
}

@inproceedings{shu2025llava,
  title={Llava-mod: Making llava tiny via moe-knowledge distillation},
  author={Shu, Fangxun and Liao, Yue and Zhang, Lei and Zhuo, Le and Xu, Chenning and Zhang, Guanghao and Shi, Haonan and Dai, Weilong and Yu, Zhelun and He, Wanggui and others},
  booktitle={ICLR},
  year={2025}
}

@article{li2026mi,
  title={{MI}-Pruner: Crossmodal Mutual Information-guided Token Pruner for Efficient MLLMs},
  author={Li, Jiameng and Tiulpin, Aleksei and Blaschko, Matthew B.},
  journal={arXiv preprint arXiv:2604.03072},
  year={2026}
}

@inproceedings{zhang2025beyond,
  title={Beyond text-visual attention: Exploiting visual cues for effective token pruning in vlms},
  author={Zhang, Qizhe and Cheng, Aosong and Lu, Ming and Zhang, Renrui and Zhuo, Zhiyong and Cao, Jiajun and Guo, Shaobo and She, Qi and Zhang, Shanghang},
  booktitle={ICCV},
  year={2025}
}

@inproceedings{yang2025visionzip,
  title={Visionzip: Longer is better but not necessary in vision language models},
  author={Yang, Senqiao and Chen, Yukang and Tian, Zhuotao and Wang, Chengyao and Li, Jingyao and Yu, Bei and Jia, Jiaya},
  booktitle={CVPR},
  year={2025}
}

@inproceedings{lin2014microsoft,
  title={Microsoft coco: Common objects in context},
  author={Lin, Tsung-Yi and Maire, Michael and Belongie, Serge and Hays, James and Perona, Pietro and Ramanan, Deva and Doll{\'a}r, Piotr and Zitnick, C Lawrence},
  booktitle={ECCV},
  year={2014},
}

@article{wang2025internvl3_5,
  title={InternVL3.5: Advancing Open-Source Multimodal Models in Versatility, Reasoning, and Efficiency},
  author={Wang, Weiyun and Gao, Zhangwei and Gu, Lixin and Pu, Hengjun and Cui, Long and Wei, Xingguang and Liu, Zhaoyang and Jing, Linglin and Ye, Shenglong and Shao, Jie and others},
  journal={arXiv preprint arXiv:2508.18265},
  year={2025}
}

@inproceedings{huang2025mixture,
  title={Mixture compressor for mixture-of-experts llms gains more},
  author={Huang, Wei and Liao, Yue and Liu, Jianhui and He, Ruifei and Tan, Haoru and Zhang, Shiming and Li, Hongsheng and Liu, Si and Qi, Xiaojuan},
  booktitle={ICLR
            },
  year={2025}
}

@article{kimiteam2025kimivl,
      title={{Kimi-VL} Technical Report}, 
      author={Kimi Team and Angang Du and Bohong Yin and Bowei Xing and Bowen Qu and Bowen Wang and Cheng Chen and Chenlin Zhang and Chenzhuang Du and Chu Wei and Congcong Wang and Dehao Zhang and Dikang Du and Dongliang Wang and Enming Yuan and Enzhe Lu and Fang Li and Flood Sung and Guangda Wei and Guokun Lai and Han Zhu and Hao Ding and Hao Hu and Hao Yang and Hao Zhang and Haoning Wu and Haotian Yao and Haoyu Lu and Heng Wang and Hongcheng Gao and Huabin Zheng and Jiaming Li and Jianlin Su and Jianzhou Wang and Jiaqi Deng and Jiezhong Qiu and Jin Xie and Jinhong Wang and Jingyuan Liu and Junjie Yan and Kun Ouyang and Liang Chen and Lin Sui and Longhui Yu and Mengfan Dong and Mengnan Dong and Nuo Xu and Pengyu Cheng and Qizheng Gu and Runjie Zhou and Shaowei Liu and Sihan Cao and Tao Yu and Tianhui Song and Tongtong Bai and Wei Song and Weiran He and Weixiao Huang and Weixin Xu and Xiaokun Yuan and Xingcheng Yao and Xingzhe Wu and Xinxing Zu and Xinyu Zhou and Xinyuan Wang and Y. Charles and Yan Zhong and Yang Li and Yangyang Hu and Yanru Chen and Yejie Wang and Yibo Liu and Yibo Miao and Yidao Qin and Yimin Chen and Yiping Bao and Yiqin Wang and Yongsheng Kang and Yuanxin Liu and Yulun Du and Yuxin Wu and Yuzhi Wang and Yuzi Yan and Zaida Zhou and Zhaowei Li and Zhejun Jiang and Zheng Zhang and Zhilin Yang and Zhiqi Huang and Zihao Huang and Zijia Zhao and Ziwei Chen}, 
  journal={arXiv preprint arXiv:2504.07491},
      year={2025},
}

@inproceedings{dao2023flashattention2,
  title={Flash{A}ttention-2: Faster Attention with Better Parallelism and Work Partitioning},
  author={Dao, Tri},
  booktitle={ICLR},
  year={2024}
}

@inproceedings{gurari2018vizwiz,
  title={Vizwiz grand challenge: Answering visual questions from blind people},
  author={Gurari, Danna and Li, Qing and Stangl, Abigale J and Guo, Anhong and Lin, Chi and Grauman, Kristen and Luo, Jiebo and Bigham, Jeffrey P},
  booktitle={CVPR},
  year={2018}
}

@inproceedings{masry2022chartqa,
  title={Chartqa: A benchmark for question answering about charts with visual and logical reasoning},
  author={Masry, Ahmed and Do, Xuan Long and Tan, Jia Qing and Joty, Shafiq and Hoque, Enamul},
  booktitle={ACL Findings},
  year={2022}
}

@inproceedings{mathew2022infovqa,
  title={Infographicvqa},
  author={Mathew, Minesh and Bagal, Viraj and Tito, Rub{\`e}n and Karatzas, Dimosthenis and Valveny, Ernest and Jawahar, CV},
  booktitle={WACV},
  year={2022},
}

@article{liu2024ocrbench,
  title={Ocrbench: on the hidden mystery of ocr in large multimodal models},
  author={Liu, Yuliang and Li, Zhang and Huang, Mingxin and Yang, Biao and Yu, Wenwen and Li, Chunyuan and Yin, Xu-Cheng and Liu, Cheng-Lin and Jin, Lianwen and Bai, Xiang},
  journal={Science China Information Sciences},
  year={2024},
}

@inproceedings{yue2023mmmu,
title={MMMU: A Massive Multi-discipline Multimodal Understanding and Reasoning Benchmark for Expert AGI},
author={Xiang Yue and Yuansheng Ni and Kai Zhang and Tianyu Zheng and Ruoqi Liu and Ge Zhang and Samuel Stevens and Dongfu Jiang and Weiming Ren and Yuxuan Sun and Cong Wei and Botao Yu and Ruibin Yuan and Renliang Sun and Ming Yin and Boyuan Zheng and Zhenzhu Yang and Yibo Liu and Wenhao Huang and Huan Sun and Yu Su and Wenhu Chen},
booktitle={CVPR},
year={2024},
}

@inproceedings{chen2024we,
  title={Are we on the right way for evaluating large vision-language models?},
  author={Chen, Lin and Li, Jinsong and Dong, Xiaoyi and Zhang, Pan and Zang, Yuhang and Chen, Zehui and Duan, Haodong and Wang, Jiaqi and Qiao, Yu and Lin, Dahua and others},
  booktitle={NeurIPS},
  year={2024}
}

@article{ai2024grok,
  title={Grok-1.5 vision preview},
  author={AI, X},
  journal={https://x. ai/blog/grok-1.5 v},
  year={2024}
}

@inproceedings{lu2024not,
  title={Not all experts are equal: Efficient expert pruning and skipping for mixture-of-experts large language models},
  author={Lu, Xudong and Liu, Qi and Xu, Yuhui and Zhou, Aojun and Huang, Siyuan and Zhang, Bo and Yan, Junchi and Li, Hongsheng},
  booktitle={ACL},
  year={2024}
}

@inproceedings{bai2025diep,
  title={DiEP: Adaptive Mixture-of-Experts Compression through Differentiable Expert Pruning},
  author={Bai, Sikai and Li, Haoxi and Zhang, Jie and Hong, Zicong and Guo, Song},
  booktitle={NeurIPS},
  year={2025}
}

@inproceedings{dong2026mmtok,
  title={Mmtok: Multimodal coverage maximization for efficient inference of vlms},
  author={Dong, Sixun and Hu, Juhua and Zhang, Mian and Yin, Ming and Fu, Yanjie and Qian, Qi},
  booktitle={ICLR},
  year={2026}
}

@article{tan2026decomoe,
  title={DecoMoE: Decoupling Visual Propagation and Expert Computation for Efficient Multimodal MoE Inference},
  author={Tan, Xudong and Ye, Peng and Xie, Ming and Huang, Chenyu and Yang, Yaoxin and Fan, Jiayuan and Chen, Tao},
  journal={arXiv preprint arXiv:2609.38823},
  year={2026}
}

@inproceedings{lu2022learn,
  title={Learn to explain: Multimodal reasoning via thought chains for science question answering},
  author={Lu, Pan and Mishra, Swaroop and Xia, Tanglin and Qiu, Liang and Chang, Kai-Wei and Zhu, Song-Chun and Tafjord, Oyvind and Clark, Peter and Kalyan, Ashwin},
  booktitle={NeurIPS},
  year={2022}
}

@inproceedings{singh2019towards,
  title={Towards vqa models that can read},
  author={Singh, Amanpreet and Natarajan, Vivek and Shah, Meet and Jiang, Yu and Chen, Xinlei and Batra, Dhruv and Parikh, Devi and Rohrbach, Marcus},
  booktitle={CVPR},
  year={2019}
}

@inproceedings{hudson2019gqa,
  title={Gqa: A new dataset for real-world visual reasoning and compositional question answering},
  author={Hudson, Drew A and Manning, Christopher D},
  booktitle={CVPR},
  year={2019}
}

@inproceedings{wang2026token,
  title={When token pruning is worse than random: Understanding visual token information in vllms},
  author={Wang, Yahong and Wu, Juncheng and Ni, Zhangkai and Yang, Longzhen and Liu, Yihang and Yang, Chengmei and Wen, Ying and He, Lianghua and Tang, Xianfeng and Liu, Hui and others},
  booktitle={CVPR},
  year={2026}
}

@inproceedings{cai2026gemoe,
  title={GeMoE: Gating Entropy is All You Need for Uncertainty-aware Adaptive Routing in MoE-based Large Vision-Language Models},
  author={Cai, Chaoxiang and Weng, Minghe and Li, Jie and Jiang, Yibo and Yang, Longrong and Qin, Zequn and Li, Xi},
  booktitle={ECCV},
  year={2026}
}

@inproceedings{li2025mbq,
  title={Mbq: Modality-balanced quantization for large vision-language models},
  author={Shiyao Li and Yingchun Hu and Xuefei Ning and Xihui Liu and Ke Hong and Xiaotao Jia and Xiuhong Li and Yaqi Yan and Pei Ran and Guohao Dai and Shengen Yan and Huazhong Yang and Yu Wang},
  booktitle={CVPR},
  year={2025}
}

@article{zheng2026tackling,
  title={Tackling Multimodal Learning Challenges with Mixture-of-Expert: A Survey},
  author={Zheng, Liangwei Nathan and Zhang, Wei Emma and Maennel, Olaf and Yue, Lin and Chen, Weitong},
  journal={arXiv preprint arXiv:2605.27431},
  year={2026}
}

@article{chen2015microsoft,
  title={Microsoft coco captions: Data collection and evaluation server},
  author={Chen, Xinlei and Fang, Hao and Lin, Tsung-Yi and Vedantam, Ramakrishna and Gupta, Saurabh and Doll{\'a}r, Piotr and Zitnick, C Lawrence},
  journal={arXiv preprint arXiv:1504.00325},
  year={2015}
}

@article{chen2026you,
  title={You Only Need 2/3 of the Chosen Experts: An Empirical Study of Dynamic Expert Pruning in Fine-Grained MoE LLMs},
  author={Chen, Yuanteng and Lai, Qiwei and Tianqi, Chen and Wang, Peisong and Shao, Yuantian and Zeng, Nanxin and Liu, Zhilei and Li, Chuangyi and Liu, Jing and Cheng, Jian},
  journal={arXiv preprint arXiv:2609.25809},
  year={2026}
}

@inproceedings{alvar2025divprune,
  title={Divprune: Diversity-based visual token pruning for large multimodal models},
  author={Alvar, Saeed Ranjbar and Singh, Gursimran and Akbari, Mohammad and Zhang, Yong},
  booktitle={CVPR},
  year={2025}
}

@article{Qwen3VL,
      title={Qwen3-VL Technical Report}, 
      author={Shuai Bai and Yuxuan Cai and Ruizhe Chen and Keqin Chen and Xionghui Chen and Zesen Cheng and Lianghao Deng and Wei Ding and Chang Gao and Chunjiang Ge and Wenbin Ge and Zhifang Guo and Qidong Huang and Jie Huang and Fei Huang and Binyuan Hui and Shutong Jiang and Zhaohai Li and Mingsheng Li and Mei Li and Kaixin Li and Zicheng Lin and Junyang Lin and Xuejing Liu and Jiawei Liu and Chenglong Liu and Yang Liu and Dayiheng Liu and Shixuan Liu and Dunjie Lu and Ruilin Luo and Chenxu Lv and Rui Men and Lingchen Meng and Xuancheng Ren and Xingzhang Ren and Sibo Song and Yuchong Sun and Jun Tang and Jianhong Tu and Jianqiang Wan and Peng Wang and Pengfei Wang and Qiuyue Wang and Yuxuan Wang and Tianbao Xie and Yiheng Xu and Haiyang Xu and Jin Xu and Zhibo Yang and Mingkun Yang and Jianxin Yang and An Yang and Bowen Yu and Fei Zhang and Hang Zhang and Xi Zhang and Bo Zheng and Humen Zhong and Jingren Zhou and Fan Zhou and Jing Zhou and Yuanzhi Zhu and Ke Zhu},
	  journal={arXiv preprint arXiv:2511.21631},
      year={2025}
}

@article{li2025minimax,
  title={Minimax-01: Scaling foundation models with lightning attention},
  author={Li, Aonian and Gong, Bangwei and Yang, Bo and Shan, Boji and Liu, Chang and Zhu, Cheng and Zhang, Chunhao and Guo, Congchao and Chen, Da and Li, Dong and others},
  journal={arXiv preprint arXiv:2501.08313},
  year={2025}
}

@inproceedings{zhang2025lmms,
  title={Lmms-eval: Reality check on the evaluation of large multimodal models},
  author={Zhang, Kaichen and Li, Bo and Zhang, Peiyuan and Pu, Fanyi and Cahyono, Joshua Adrian and Hu, Kairui and Liu, Shuai and Zhang, Yuanhan and Yang, Jingkang and Li, Chunyuan and others},
  booktitle={NAACL Findings},
  year={2025}
}

@inproceedings{yang2025vflowopt,
  title={Vflowopt: A token pruning framework for lmms with visual information flow-guided optimization},
  author={Yang, Sihan and Xu, Runsen and Cui, Chenhang and Wang, Tai and Lin, Dahua and Pang, Jiangmiao},
  booktitle={ICCV},
  year={2025}
}

@inproceedings{zhang2025cdpruner,
  title={Beyond Attention or Similarity: Maximizing Conditional Diversity for Token Pruning in MLLMs},
  author={Zhang, Qizhe and Liu, Mengzhen and Li, Lichen and Lu, Ming and Zhang, Yuan and Pan, Junwen and She, Qi and Zhang, Shanghang},
  booktitle={NeurIPS},
  year={2025},
}

@inproceedings{wen2025stop,
  title     = {Stop looking for important tokens in multimodal language models: Duplication matters more},
  author    = {Wen, Zichen and Gao, Yifeng and Wang, Shaobo and Zhang, Junyuan and Zhang, Qintong and Li, Weijia and He, Conghui and Zhang, Linfeng},
  booktitle = {EMNLP},
  year      = {2025},
}
}

\end{document}